\documentclass[10pt,twocolumn,letterpaper]{article}

\usepackage[pagenumbers]{cvpr}
\usepackage{times}
\usepackage{epsfig}
\usepackage{graphicx}
\usepackage{amsmath}
\usepackage{amssymb}
\usepackage{booktabs}

\usepackage{hyperref}
\hypersetup{colorlinks,allcolors=black}

\usepackage[capitalize]{cleveref}
\crefname{section}{Sec.}{Secs.}
\Crefname{section}{Section}{Sections}
\Crefname{table}{Table}{Tables}
\crefname{table}{Tab.}{Tabs.}

\begin{document}

\title{Optimization-Based Eye Tracking using Deflectometric Information}

\author{\textbf{Tianfu Wang$^1$, Jiazhang Wang$^2$, Oliver Cossairt$^2$, Florian Willomitzer$^3$}\\ \\
$^1$Department of Computer Science, ETH Zürich, Zürich, Switzerland, 8092\\%
$^2$Department of Electrical and Computer Engineering, Northwestern University, Evanston, IL, 60208\\%
$^3$Wyant College of Optical Sciences, University of Arizona, Tuscon, AZ, 85721
}
\maketitle

\begin{abstract}
Eye tracking is an important tool with a wide range of applications in Virtual, Augmented, and Mixed Reality (VR/AR/MR) technologies. State-of-the-art eye tracking methods are either reflection-based and track reflections of sparse point light sources, or image-based and exploit 2D features of the acquired eye image. In this work, we attempt to significantly improve reflection-based methods by utilizing pixel-dense deflectometric surface measurements in combination with optimization-based inverse rendering algorithms. Utilizing the known geometry of our deflectometric setup, we develop a differentiable rendering pipeline based on PyTorch3D that simulates a virtual eye under screen illumination. Eventually, we exploit the image-screen-correspondence information from the captured measurements to find the eye's \textit{rotation}, \textit{translation}, and \textit{shape} parameters with our renderer via gradient descent. In general, our method does not require a specific pattern and can  work with ordinary video frames of the main VR/AR/MR screen itself.  We demonstrate real-world experiments with evaluated mean relative gaze errors below $0.45 ^{\circ}$ at a precision better than $0.11 ^{\circ}$. Moreover, we show an improvement of 6X over a representative reflection-based state-of-the-art method in simulation.
\end{abstract}

\begin{figure*}[htp]
    \centering
    \includegraphics[width=1.0\textwidth]{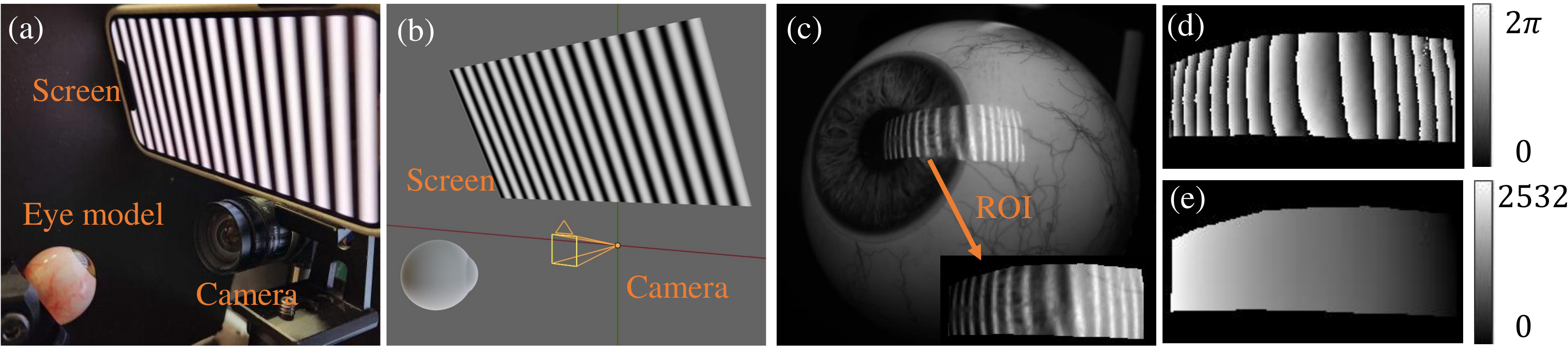}
    \caption{\textbf{Setup and retrieved measurements.} (a) Real world experimental setup. A mobile phone is used as the screen. The physical eye model is mounted on a rotation stage and observed by a camera (b) Simulated setup, used in our inverse rendering pipeline (c) Captured image of the physical eye model, illuminated with a high frequency (16 periods) vertical sinusoidal pattern. (d) Wrapped phase map retrieved from phase shifting method. (e) Unwrapped phase map containing image-screen correspondence information (here for vertical direction).}.
    \label{fig:setup}
\end{figure*}

\section{Introduction}
The seemingly infinite possibilities that Virtual, Augmented, and Mixed Reality (VR/AR/MR) technologies have to offer kindled immense excitement in the graphics and vision communities. However, as of today, a variety of unsolved problems still impede the widespread adoption of VR/AR/MR technologies in consumer and industrial spheres. One of these unsolved problems is fast and accurate eye tracking. If solved, 
accurate and fast tracking of a user's eye position and gazing direction  will have a significant impact on the advancement of VR/AR/MR technologies - first and foremost because it is the missing puzzle piece to enable resource-efficient foveated rendering techniques that result in better frame rates and higher (effective) resolution \cite{patney2016towards}. Moreover, accurate and fast eye tracking allows for the calculation of the viewer's focus distance by monitoring the inter-pupillary distance  and thus improving the viewing comfort \cite{kim2019effects}, and it can help with the creation of more realistic virtual human avatars \cite{mikhailenko2022eye}. 
Amongst the different possibilities to track the gaze of a human eye \cite{khaldi2020laser, barea2002system}, optical methods prove to be particularly advantageous, because they work fast and contact-less. Many ``active'' optical methods do also not require external markers on the eye or the surrounding periocular region, which makes them highly desirable for consumer eye tracking solutions. Current optical marker-free eye tracking methods can be broadly categorized into two groups: Image-based methods and reflection-based methods \cite{chamberlain2007eye, hansen2009eye}.  

Image-based methods retrieve geometrical information about the gaze direction from 2D features in captured 2D images of the eye \cite{krafka2016eye, wang2017real, angelopoulos2020event, tonsen2017invisibleeye}. Pupil, iris, limbus, or eyelids, are popular candidates for 2D features  to calculate eye position and gaze direction. Although some techniques flood-illuminate the eye  for better visibility (e.g., with infrared light) \cite{1187975, morimoto2005eye}, those methods can be considered as ``passive'' methods since the illumination is not actively spatially modulated. Traditionally, classic computer vision techniques, such as edge detection and model fitting, are utilized to find the 2D features in the eye images. Recently, machine learning and deep learning techniques have been employed to achieve better feature extraction quality \cite{tonsen2017invisibleeye}. However, image-based approaches still rely solely on 2D image space information, although  explicit 3D information about the eye can lead to better and more robust tracking. Moreover, the density of extracted 2D features is still relatively sparse, which impedes the reconstruction quality. Lastly, most image-based models fail to capture the light transport complexities of the lens surface of the eye. The refraction of light at transparent eye parts like iris or pupil can cause view-dependent deviations in apparent feature locations \cite{patel2019refractive}, making it difficult to model these optical complications using a purely image-based approach. To circumvent this problem, current approaches often resort to rely on unreliable secondary features, such as face and skin movement \cite{wang2017real}, to predict the eye gazing directions.

The second class of methods exploits the (partial) specularity of the eye surface to capture real 3D information that is then used to calculate the gazing direction. A prominent example for such ``reflection-based methods''  is ``glint-tracking'': the reflections of a few sparse (infrared) point light sources is  observed with a camera over the reflective cornea  surface \cite{chamberlain2007eye, hansen2009eye}. The position of the corneal reflections in the camera image changes depending on the rotation and translation of the eye. These changes in position are used in conjunction with other eye features, such as the pupil position and geometry, to evaluate the gaze direction. Over the years, the point light source arrangements have evolved from single point lights (or ``glints'') with one camera \cite{guestrin2006general}, to multi-view
\cite{hennessey2006single, 1211502}, and multi-point-source \cite{mestre2018robust, hennessey2009improving} setups, with more sampled surface points generally leading to higher accuracy in gaze evaluation
\cite{mestre2018robust}. However, although the number of reflection points that sample the eye surface has increased, state-of-the-art methods still use only roughly $\sim$12 reflection points. Compared to the number of pixels in the camera image space, this is still very \textit{sparse}, and hence only \textit{sparse} surface information from the eye can be extracted. \\

In this paper, we introduce a simple but effective method to \textit{drastically increase} the number of reflection points and thus to obtain a \textit{pixel-dense sampling} of the eye surface: We replace the array of sparse point light sources with an extended self-illuminated screen. The screen displays a specific (known) pattern, and the reflection of this pattern over the eye surface is observed with the camera (Fig.~\ref{fig:setup}c). The deformation of the pattern in the camera image reveals information about the rotation and translation of the measured eye as well as its surface (if desired). We emphasize that each pixel in our used screen plays the role of an individual point light source. For example, the usage of an HD display ($1920 \times 1080$ pixels) yields to $> 2 Mio$ point light sources, which, in comparison to a 12-point sources setup, increases the number of light sources by a factor $>170,000$. Although we will show later that geometrical and perspective constraints prevent us from exploiting the information from each of the  $> 2 Mio$ pixels, it can be easily understood that our method leads to a drastic increase in acquired information about the eye surface that can be exploited to evaluate the gaze direction.

We note that the  basic idea to use an extended screen for the dense 3D measurement of specular surfaces has been long known in the metrology community, where the related method is referred to as ``deflectometry'' \cite{knauer2004phase, huang2018review, hausler2013deflectometry, faber2012deflectometry}. Traditionally, deflectometry is used for 3D surface inspection tasks, such as quality assessment of lens surfaces \cite{knauer2008measuring}, car bodies \cite{hofer2016infrared}, free-form surfaces \cite{rottinger2011deflectometry}, with recent applications in cultural heritage analysis \cite{willomitzer2020hand, xu20193d, li2021low} or also as ``differentiable refractive deflectometry'' \cite{wang2021towards}. Measurements of the human cornea to quantify vision impairments have been shown as well \cite{liang2016single}. An approach that uses ``classical'' deflectometry for eye tracking has been recently published in \cite{wang2022easy, wang2021vr}: the  eye's surface shape and normal map is measured via deflectometry and these 3D measurements are used to extract 3D features that reveal the gaze direction.

In our contribution, we propose to exploit the densely captured deflectometric information of the eye surface (i.e., the screen reflection observed over the eye surface) as well, but employ an \textit{inverse-rendering procedure} to evaluate the eye's rotation and translation parameters: We simulate a realistic computer-generated (CG) eye model that we place in a virtual copy of our (calibrated) experimental setup. By comparing the rendered images of the CG eye with our real measurements, we can optimize the eye's rotation and translation through gradient descent. Simultaneously, our differentiable renderer also allows for joint gradient descent optimization of gaze direction \textit{and} CG eye shape, leading to an experimentally evaluated mean relative gaze error below $0.45^\circ$.
In comparison to \cite{wang2021vr, wang2022easy}, our introduced approach works with only one camera. Moreover, we can facilitate any possible pattern/image that is displayed on the screen. In the end, this completely removes the need for a second ``pattern-screen'' in the VR/AR/MR headset. We simply use the main screen of the headset and it's actual displayed content (movie, video game stream, etc.) to jointly evaluate the gaze direction and eye shape. This process can be performed solely from a single-shot capture, allowing for motion robust eye tracking solutions with the lowest possible latency.
We summarize our contributions as follows:

\begin{itemize}
  \setlength\itemsep{-0.2em}

\item We introduce a novel approach that uses dense deflectometric information together with an optimization-based framework that leverages differentiable rendering to determine the eye's translation and rotation parameters.

\item We develop a novel differentiable deflectometry shader based on the PyTorch3D renderer \cite{ravi2020accelerating} that simulates the light transport from a pattern screen illumination on a specular object, such as the eye.

\item We extend our framework to jointly optimize gaze direction and the shape of the CG eye model. For real-world experiments, this allows for a more realistic representation of the real eye, which in turn allows better evaluation results and additional potential features, such as in-headset automatic vision correction.
\item We introduce different flavors of our approach (including one that uses ordinary video frames of the main VR/AR/MR screen). We demonstrate real-world experiments with evaluated precisions in gaze estimation between $0.11^\circ$ and only $0.02^\circ$ and relative gaze estimation errors between $0.45^\circ$ and only $0.27^\circ$. Moreover, perform a quantitative comparison of our different flavors with a representative state-of-the-art method \cite{mestre2018robust} and show that our method performs up to 6X better in both mean and standard deviation of the gaze error.

\end{itemize}

\section{Methods}

\subsection{Gaze estimation via differential rendering}
With a calibrated screen-camera setup, we capture one image (or potentially more images) of the eye with screen reflection,  denoted as $\{I^{gt}\}$ (Fig.~\ref{fig:setup}c). Our goal is to estimate the \textit{rotation}, \textit{translation}, and \textit{shape} of the eye. We approach this goal by formulating a joint optimization problem of the eye parameters. We define the true rotation, translation, and shape of the eye during each measurement as the parameter set $v$. The goal of our inverse rendering optimization procedure is to find the optimal parameter set $v^*$ using a base CG eye model (see Fig.~\ref{fig:render}) and the observed view of the camera $\{I^{gt}\}$. This optimization is performed by minimizing a pre-defined objective function, denoted as $\mathcal{L}$: 

\begin{equation}
    v^* = argmin \  \mathcal{L}(I(v), v; \{I^{gt}\})
\end{equation}

We minimize our objective using gradient descent. The function $I$ represents the differentiable rendering function that takes in the eye parameters $v$. Our differentiable rendering module is specified in Sec.~\ref{dif}, and our eye parameter representation and optimization strategy is detailed in Sec.~\ref{shape} and the Suppl. Material. We note that the captured eye image $\{I^{gt}\}$ is the crucial input of our algorithm, as our loss function $\mathcal{L}$ is dependent on the information we extract from the captured eye image. In the following sections, we will introduce two different differential rendering-based eye-tracking procedures from deflectometric measurements: one extracting 3D screen-camera image correspondence information decoded from the deformed patterns in the captured image(s) (Sec.~\ref{corr}), and another procedure leveraging directly the intensity values of the captured image and calculating the ``photometric loss'' (Sec.~\ref{col}).

Our setup for capturing the screen-illuminated eye images consists of a camera and a rectangular screen that displays a pattern (e.g., a sinusoid). An image of our experimental setup is shown in Fig.~\ref{fig:setup}a, and Fig.~\ref{fig:setup}b shows an image of our simulated setup that we use to develop our differential rendering algorithms and to compare our method with other techniques (see Sec.~\ref{comp}). In our real-world experiment, our measurement object is a realistic physical 3D eye model that is mounted on a rotation stage. We emphasize again that screen and camera in our setup are calibrated, i.e., the intrinsic camera parameters as well as the position of the screen relative to the camera is known within the bounds of the calibration error.

\subsection{Modeling eye anatomy and geometry}

\begin{figure}[t]
  \centering
   \includegraphics[width=1\linewidth]{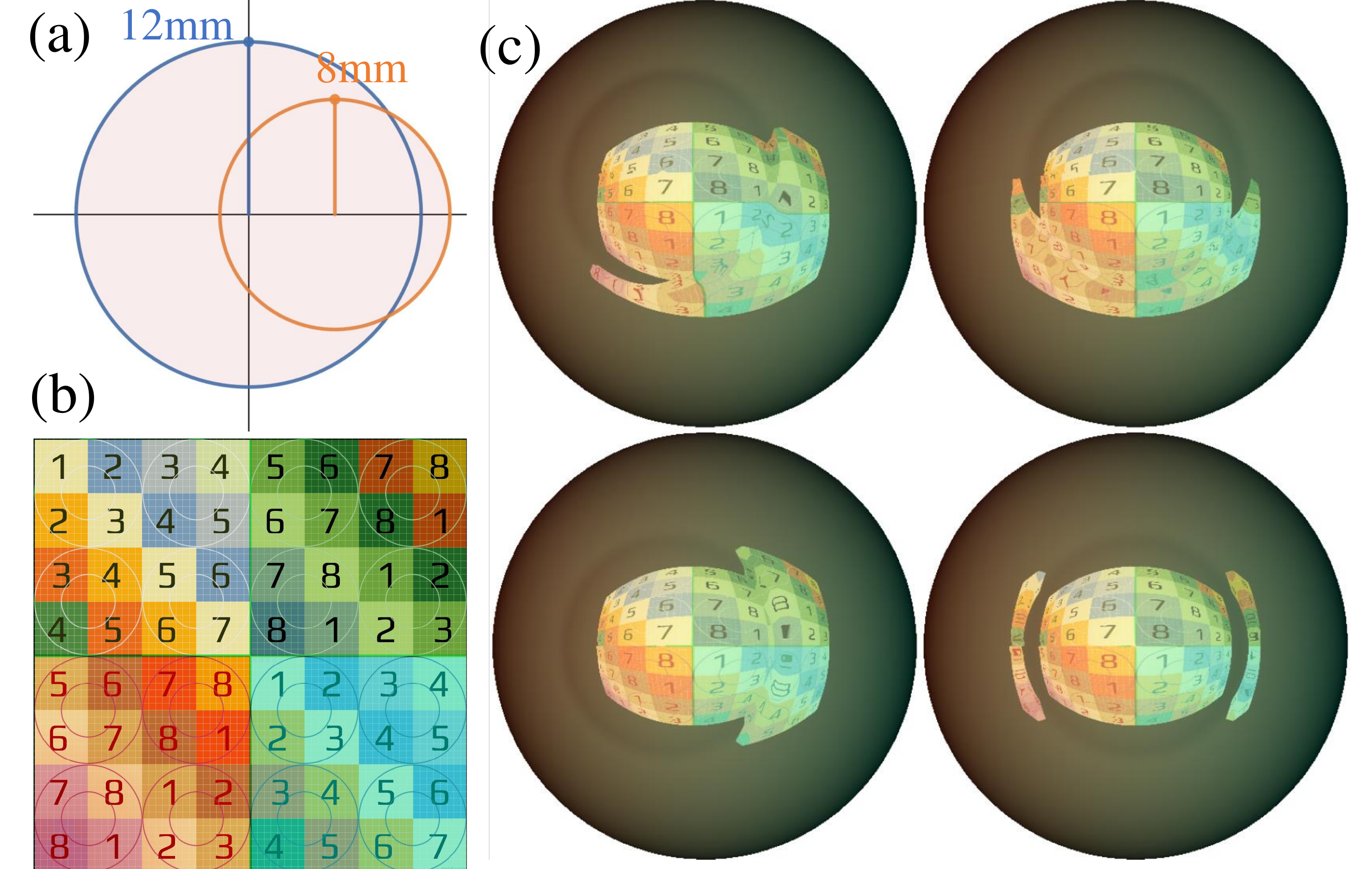}
   \caption{\textbf{Base CG eye shape and example for screen illumination renders.} (a) The surface of the base CG eye model is formed by two intersecting spheres of 12mm radius for the sclera and 8mm radius for the cornea. The centers of the two spheres have a distance of approximately 6mm. (b) Example for illumination pattern (color coded). (c) Respective renders of the CG eye model for different gaze directions.}
   \label{fig:render}
\end{figure}

The human eye is a highly complex organ that has been the subject of extensive research and analysis with regards to its anatomy and geometry \cite{remington2021clinical, willoughby2010anatomy}. Our eye tracking approach relies on a virtual CG model of the human eye. Hence, anatomic knowledge is crucial for improving the quality of gaze estimation algorithms. In this contribution, we concentrate our modeling efforts on a realistic representation of the eye \textit{surface}, in particular, the region close around the limbus - the transition between cornea and sclera. It has been shown in \cite{wang2021vr} that, due to the prominent surface feature at the limbus, this region is especially suitable to extract information about the gazing direction. We create our CG eye base model on the assumption of two intersecting spheres (the surfaces of sclera and cornea), with the sclera-sphere having a larger radius of 12mm and the cornea-shere having a smaller radius of 8mm. The distance of both sphere centers is approximately 6mm (see Fig.~\ref{fig:render}a).  We emphasize that these measures have been obtained from statistical size metrics literature \cite{remington2021clinical, willoughby2010anatomy} and are estimated to provide a fairly accurate base CG eye model. We will discuss in  Sec.~\ref{shape} how we employ iterative shape optimization on the  base CG eye model to approach an even more realistic eye shape and to possibly compensate for individual deviations of the ``common'' eye shape, e.g., for cornea deformations. 

\subsection{Differentiable Deflectometry Rendering}
\label{dif}
The differentiable deflectometry rendering function is crucial in allowing the eye parameters to properly update towards low error gaze estimation. We utilize  the PyTorch3D framework \cite{ravi2020accelerating} which is a rasterizer-based differentiable renderer that provides the necessary tools to perform differentiable transformations from virtual world space to virtual image space using a perspective camera model. The framework also allows to find the closest intersection between a camera ray and the mesh geometry, and provides us with object surface depth and normal information for each pixel or our rendered virtual object. However, native PyTorch3D does not support indirect or area lighting calculations. To overcome this limitation, we designed a specialized deflectometry shader similiar to \cite{9443521}, but tailored to our specific eye tracking task. Our shader simulates specular reflection from area lighting and acts as a single bounce ray-tracer that calculates the mesh position and object surface normal for each camera image pixel using the PyTorch3D rasterizer. The view direction is calculated as a vector originating from each camera pixel and pointing toward the mesh position. The specular reflection ray is then computed by reflecting each view direction vector at the surface of the mesh with the surface normal obtained from the PyTorch3D rasterizer. Intersecting the specular reflection ray with the screen delivers the (color) intensity value at the respective pixel and establishes correspondence between simulated screen and simulated camera. We provide example renderings of our base CG eye model under multiple gaze angles in Fig.~\ref{fig:render}c, using a color-coded screen pattern (Fig.~\ref{fig:render}a) for demonstration purposes. We note that the diffuse component is calculated as well and acts as a background in our rendered images(see  Fig.~\ref{fig:render}c). By implementing our lighting calculation as a PyTorch module and utilizing the differentiable rasterizer in PyTorch3D, we have created a completely differentiable rendering pipeline for deflectometric eye measurements that enables us to optimize scene parameters, such as the rotation and translation of the mesh geometry as well as the mesh itself, through gradient descent.

\subsection{Eye Tracking Using Deflectometric Correspondences}
\label{corr}

 We now go into detail on gaze estimation using screen-image correspondence information. We obtain correspondence using standard procedures of optical 3D imaging as found, e.g., in deflectometry \cite{knauer2004phase,huang2018review, wang2021mitsuba}, or active triangulation (structured light) \cite{srinivasan1984automated, schaffer2010high, willomitzer2017single} literature: Structured patterns are displayed on the screen and reflected over the eye surface. Eventually  deformed pattern in the camera image is decoded to calculate a correspondence between screen pixels and camera pixels. In our experiments, we utilize various patterns to extract this correspondence information, balancing the density of correspondence points and the number of shots. For instance, we can display a single checkerboard pattern and utilize checkerboard corner detection algorithms to extract correspondence points from the corners of the pattern in single-shot. We also can display sinusoidal fringe patterns and obtain pixel-dense correspondence via, e.g.,  the  four phase-shift algorithm \cite{servin2009general} and phase unwrapping -- however at the cost of multiple sequentially captured images. For the mentioned arbitrary image content of movie video frames or video game streams in VR/AR/MR, we can use SIFT feature matching \cite{lowe2004distinctive} to extract correspondence information between the display image and its distorted reflection on the eye. Other patterns/images and correspondence evaluation methods are possible as well. Related experiments and quantitative comparisons are shown in Sec.~\ref{comp}. 
 
\begin{figure}[t]
  \centering
   \includegraphics[width=1\linewidth]{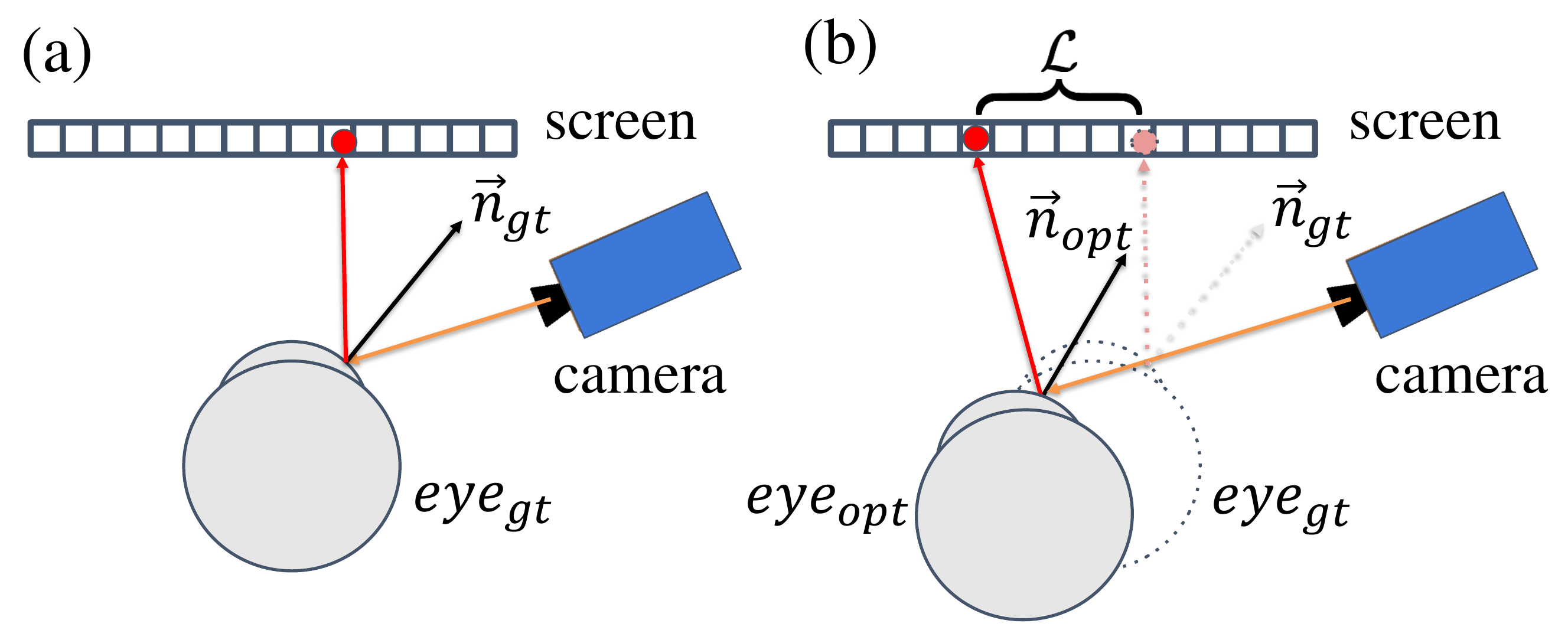}
   \caption{\textbf{Eye Tracking Using Deflectometric Correspondences.} (a) We illuminate the eye using sinusoidal patterns, and obtain pixel-dense screen-image correspondence points. (b)~Our differentiable renderer then simulates a virtual deflectometric setup and attempts to find the correct eye parameter by minimizing screen correspondence point distances between simulation and captured images.}
   \label{fig:def_cor}
\end{figure}

 After establishing correspondence between screen pixels and camera pixels, we represent the set of image pixel positions as $\mathcal{P}_{corr} = \{p\}$, and the corresponding  sub-pixel precise location on the screen as $\{corr(p)\}$. The set of correspondences ${(p, corr(p))}$ is  the input to our differentiable rendering optimization pipeline. Our differentiable renderer takes ${(p, corr(p))}$ and attempts to optimize a CG eye model in a virtual deflectometry setup to fit the measured correspondence set with the simulated correspondence set (see Fig.~\ref{fig:def_cor}) as follows: In our virtual model, we trace rays through the same set of virtual image pixels $\mathcal{P}_{corr}$. By simulating the specular reflection light transport at our CG eye model surface, we obtain the intersection location between virtual reflected rays and virtual screen plane, denoted as $\{corr_{opt}(p)\}$. The difference between $\{corr(p)\}$ and $\{corr_{opt}(p)\}$ is dependent on the base CG eye model's shape, rotation, and  translation in virtual space.

 We normalize the screen correspondence points to a 2D planar coordinate system so that the point coordinates that directly hit the screen have values between -1 and 1. For our given image pixel positions $\mathcal{P}_{corr}$, we can now calculate the squared distance between the measured correspondence points $\{corr(p)\}$ and the rendered screen-plane intersection points $\{corr_{opt}(p)\}$. Our optimization objective is then to minimize the mean of the log of this squared distance for all obtained correspondences.

\begin{subequations}
\begin{equation}
  \mathcal{L} = \frac{1}{|\mathcal{P}_{corr}|} \sum_{p\in \mathcal{P}_{corr}} dist(||corr(p) - corr_{opt}(p)||^2),
  \label{eq:important}
\end{equation}
where we have: 
\begin{equation}
dist(d) =\begin{cases} 
      d & d \leq 1 \\
      \log (d) + 1 & d >1 \\
   \end{cases}
\end{equation}
\end{subequations}
Note that the log is only applied in the rare cases when the distance  between the correspondence points is larger than 1 (half of the screen dimension) and would cause large values in the loss function.

\subsection{Eye Tracking Using a Photometric Loss}
\label{col}

In the previous section we introduced our pipeline that uses screen-camera image correspondence information to find the eye gaze direction. We mentioned that, e.g., SIFT features \cite{lowe2004distinctive} can be used to extract correspondence information for arbitrary images. However, the VR/AR/MR content might lack large numbers of features in certain situations, e.g., if texture-less low-frequency images (sky, water, etc.) are displayed. Another possibility is that the displayed features are too distorted in the camera image (see Fig.~\ref{fig:render}c) and cannot be detected properly.
In theses cases,  the previously described correspondence method becomes unreliable. For this reason, we implemented a second flavor of our method that optimizes rotation, translation, and shape of the CG eye model based on ``photometric loss'', i.e., by directly comparing the intensity values in each camera pixel ($I_{gt}(p)$) with the intensity values in the simulated image ($I_{opt}(p)$). In our differentiable renderer pipeline, we now use the objective of the photometric loss, which is the mean absolute per-pixel RGB intensity difference between the captured camera image and the optimized renders.

\begin{equation}
  \mathcal{L} = \frac{1}{|\mathcal{P}|} \sum_{p\in \mathcal{P}}|I_{gt}(p) - I_{opt}(p)|
  \label{eq:important}
\end{equation}

It will be shown in Sec.~\ref{comp} that, in general, the photometric loss method delivers a slightly higher gazing estimation error than the  correspondence method. For this reason, it should be seen more as a ``backup-addition'' to the correspondence method for the cases where no/not enough correspondences can be found.

\subsection{Optimizing the eye shape}
 \label{shape}

Our method uses a differentiable renderer to simulate deflectometric images of a  CG eye model in a virtual scene, where the eye model is moved/rotated in the virtual space based on the gradient descent optimization. This means that our method heavily relies on a realistic shape of the used eye model. For real-world experiments it is possible that the shape of the measured eye is different for different subjects, e.g., if a user has corneal deformations. For an improved robustness of our technique, it is therefore necessary to develop additional methods to accommodate for varying eye shapes.

We propose to take advantage of our differentiable rendering framework and jointly optimize for the shape of the virtual eye model along with its translation and rotation. One typical method to perform shape optimization is to directly optimize for the position of the vertices of the mesh, or optimize a per-vertex displacement field on top of a base mesh \cite{worchel}. However, we found that directly optimizing on the whole mesh introduces a very high dimensional optimization space that often leads to local minima. For this reason, we add an additional constraint on the eye shape that drastically reduces the optimization space: We assume that the eye is rotational symmetric around its optical axis. Under this assumption, we can model the shape of the eye as a set of connected circular edge loops, all centered around the optical axis. Our shape consists of the vertices $V \in \mathbb{R}^{H X N}$, where $V$ is the number of of edge loops, and $H$ is the number of vertices for each edge loop. Setting the axis of rotation of the eye model to the local Z axis, and the center of the (initially spherical) sclera  at the origin, we can define $c_0 < c_1 < ...< c_{N -1}$ as the z coordinates of the center of the $N$ edge loops. If we then define the radii of the $N$ edges loops as $r_0, r_1, ..., r_{N - 1}$, then the 3D local coordinate of the $i^{th}$ vertex of the $j^{th}$ vertex loop can be written as $(r_j \cos(2\pi i /H), r_j \sin(2\pi i/H, c_j)$. 

\begin{figure}[t]
  \centering
   \includegraphics[width=1.0\linewidth]{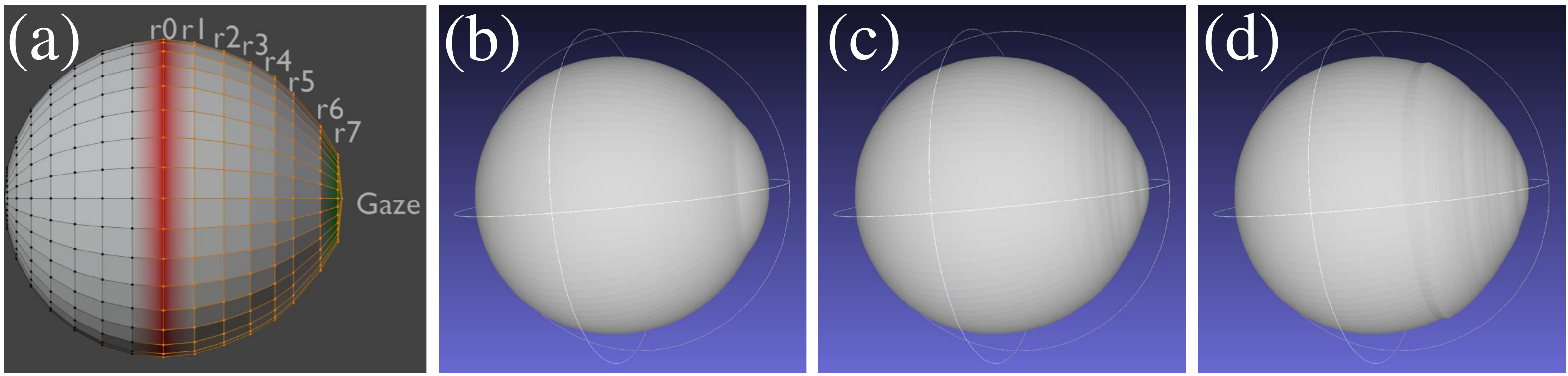}
   \caption{\textbf{Eye shape optimization.} (a) Side view of the low polygon version of the eye model ($N = 8$). We show the circular edge loops (yellow) that we optimize for and their corresponding radii (the red loop for $r_0$, the green loop $r_7$). (b) base eye shape. (c,d)~Optimized eye shape with (c) and without regularizers~(d)}
   \label{fig:loops}
\end{figure}

During our shape optimization, we optimize the radii of the edge loops $R = \{r_0, r_1, ..., r_{N - 1}\}$ for the frontal part of the eye model. Figure~\ref{fig:loops}a visualizes the basic idea and geometry for N = 8 edge loops. Results are shown in Fig.~\ref{fig:loops} c and d. For our experiments we chose the number of edge loops and the number of vertices within each edge loop both to be 100.

Although this procedure greatly simplifies the shape optimization problem, a joint optimization of radii along with the translation and rotation could easily lead to local minima which results in wrong shapes (see Fig.~\ref{fig:loops}d). To address this, we utilize additional regularizers to the eye geometry to favor smooth shapes without undesirable bumps. This step is inspired by previous works that involve the use of differentiable rendering for inverse problems \cite{luan2021unified}, and regularize both the radii gradient $\mathcal{L}_{geom}$, the local radii smoothness $\mathcal{L}_{mc}(R)$ and the Laplacian of the mesh $M$, $\mathcal{L}_{lap}(M)$. These regualrizers are controlled by hyper-parameters $\lambda_{grad}, \lambda_{mc}, \lambda_{lap}$

\begin{equation}
\begin{aligned}
\mathcal{L}_{geom}(R, M) = \lambda_{grad}\mathcal{L}_{grad}(R)  + \lambda_{mc}\mathcal{L}_{mc}(R) \\ + 
\lambda_{lap}\mathcal{L}_{lap}(M)
\end{aligned}
\end{equation}

\noindent
We first regularize the gradient of the radii so that the radii form a monotonically decreasing sequence from the circle closest to the center to the furthest, which is in accordance with the anatomy of a real eye. 

\begin{equation}\label{eq:grad}\mathcal{L}_{grad}(R) = \sum_{i} \max(0, r_i - r_{i + 1})\end{equation}

\noindent
Second, we regularize the smoothness of the radii. This is to prevent the eye surface from being uneven.  Our smoothness formulation can be written as

\begin{equation}\label{eq:mc}\mathcal{L}_{mc}(R) = \sum_{i} max(MC(r_i, r_{i + 1}, r_{i + 2})- t_{mc}, 0)\end{equation}

\noindent
Here $MC$ is the discrete Menger curvature \cite{leger1999menger} of three points, where a larger Menger curvature means a more curved surface. In other words, we apply a threshold $t_{mc}$ on the local curvature of the surface.

Lastly, we add the mesh Laplacian as a regularizer to further enforce smoothness of the optimized mesh. 

\begin{equation}\label{eq:lap} \mathcal{L}_{lap}(M) =||\mathrm{L}\mathrm{V}||^2\end{equation}

\noindent
$\mathrm{V}$ is the vector of vertices in the mesh and $\mathrm{L}$ is the laplacian matrix of the mesh \cite{nealen2006laplacian}. In our experiments, we typically set $\lambda_{grad} = 0.05,\lambda_{lap} = 0.1$. For Menger curvature smoothness, we typically set $\lambda_{mc} = 0.1,t_{mc} = 4$.

We emphasize again that our proposed shape optimization not only leads to a more precise gaze estimation, but also delivers a shape of the eye model that is much closer to the real eye shape of a subject. Moreover, the shape optimization will deliver different eye shape results for different subjects. We propose that a future version of our shape optimization algorithm could be potentially used in future works to accurately \textit{measure} the eye surface during eye tracking. This would allow, e.g., for the automatic correction of vision impairments and could potentially lead to ``self-correcting'' VR/AR/MR headsets.

\section{Results}

\subsection{Real-world experiments}
\label{real_exp}
To  validate our joint shape and gaze optimization model in a quantitative fashion, we conduct real-world experiments on a realistic physical 3D eye model that emulates the shape and reflective properties of a human eye. Our experimental setup (including physical eye model) is shown in Fig.~\ref{fig:setup}a, and a closeup view of the physical eye model is shown in Fig.~\ref{fig:setup}c. We use an \textit{iPhone 13 Pro} as screen ($1170 \times 2532$ pixels) and a \textit{FLIR~fl3-u3-13s2c} as camera. The distance of the camera to our eye model is approximately 8cm, meaning that the prototype setup is already very compact.

As discussed, our algorithm does not know the shape of the physical eye model in advance, only the calibrated camera and screen position. Since the \textit{absolute} ground truth gaze direction of the eye model cannot be evaluated, we propose to use \textit{relative gazing angles} for our quantitative error evaluation: we centered the 3D eye model on a rotation stage and rotated the 3D eye model multiple times to $-4^\circ$, $-2^\circ$, $0^\circ$, $2^\circ$, and $4^\circ$. At each rotation position, we took a measurement of the 3D eye model and moved to the next rotation position.  We took 20 measurements at each of the 5 rotation positions, i.e. 100 measurements in total. We emphasize that we \textit{always} rotated the 3D model before we took a measurement, meaning that we never took two consecutive measurements at the same rotation position. In this experiment our screen displayed a  phase-shifted sinusoidal pattern in the horizontal and vertical direction, respectively. The used sinusoidal pattern had 16 periods in the horizontal direction, and (according to the screen aspect ratio) 7.4 periods in the vertical direction. The acquired phasemaps have been unwrapped with MATLAB's {\texttt{unwrap()}} function, which works sufficiently well for low noise levels and smooth surfaces.

For each of the 5 rotation positions $a$, we calculate the \textit{precision} $\sigma_r$ of our measurements, which is defined as the standard deviation of the 20 acquired measurements:

\begin{equation}\label{eq:sig}
    \sigma_a = \sqrt{\frac{\sum^{n}_i(\theta_{a, i} -  \bar{\theta_a})^2}{n}}
\end{equation}

Here, $\bar{\theta_a}$ is the mean evaluated gazing angle at each rotation position $a$, $\theta_{a, i}$ is the  $i^{th}$ measurement at rotation position $a$, and $n = 20$ in this experiment. Moreover, we define the \textit{mean relative error}  $\epsilon_0$ of the gazing direction at each rotation position $a$ with respect to $a=0^\circ$ as

\begin{equation}\label{eq:eps}
    \epsilon_0= |\bar{\theta_{a}} - \bar{\theta_{0}}|
\end{equation}

We note that, due to the definition of our mean relative error, the values with respect to other rotation positions $a$ are slightly different. A table containing $\epsilon_{-4}, \epsilon_{-2}$, $\epsilon_{2}$, and $\epsilon_{4}$ for all rotation  positions can be found in the Suppl. Material.
The results of our experiment are shown in Tab.~\ref{tab: real_exp_data}. Both datasets use the CG eye model shape optimization via radial loop optimization. If we apply the radial loop optimization \textit{together} with the additional regularizers of Eq.~\ref{eq:grad}, \ref{eq:mc}, and \ref{eq:lap}, we achieve precision values between $0.11^\circ$ and only $0.02^\circ$ and a mean relative error $\epsilon_0$ with respect to $a= 0^\circ$ between $0.45^\circ$ and only $0.27^\circ$. This demonstrates the robustness of our joint shape and gaze optimization model for real world experiments.

We additionally conduct an ablation study on the effect of our shape regularizers (also shown in Tab.~\ref{tab: real_exp_data}): If we only use the radial loop optimization \textit{without} the additional regularizers of Eq.~\ref{eq:grad}, \ref{eq:mc}, and \ref{eq:lap}, we generally achieve a slightly worse mean relative error and comparable values for the precision. This indicates that the shape regularizers have little effect on the statistical variance of the evaluated results, but help to converge to a result with overall smaller estimation error.

\begin{table}[!h]
\center
\caption{\textbf{Gaze angle estimation for real-world experiments.}}
\label{tab: real_exp_data}
\scriptsize
\begin{tabular}{l|r|r|r|r|r}\toprule
\textbf{Rotation position} &\textbf{0$^\circ$ } &\textbf{2$^\circ$} &\textbf{4$^\circ$} &\textbf{-2$^\circ$} &\textbf{-4$^\circ$} \\\midrule
Precision $\sigma_a$  &0.02$^\circ$ &0.10$^\circ$ &0.08$^\circ$ &0.11$^\circ$ &0.11$^\circ$ \\
Mean relative error $\epsilon_0$ &0$^\circ$ &0.33$^\circ$ &0.28$^\circ$ &0.27$^\circ$ &0.45$^\circ$ \\
\midrule
$\sigma_r$ w/o shape regularizer &0.04$^\circ$ &0.02$^\circ$ &0.05$^\circ$ &0.09$^\circ$ &0.06$^\circ$ \\
$\epsilon_0$ w/o shape regularizer &0$^\circ$ &0.01$^\circ$ &0.49$^\circ$ &0.44$^\circ$ &1.17$^\circ$ \\

\bottomrule
\end{tabular}
\end{table}

We conclude this section by noting that we conducted the experiment above with a rotation stage that was rotated by hand. This might have imparted an additional angular variance on the measurements that propagates through our evaluations to our results. The usage of a high-precision automated rotation stage potentially yields even better results.

\subsection{Comparison of our proposed approaches to Glint-based Eye Tracking}
\label{comp}

\begin{figure*}[t!]
    \centering
    \includegraphics[width=0.8\linewidth]{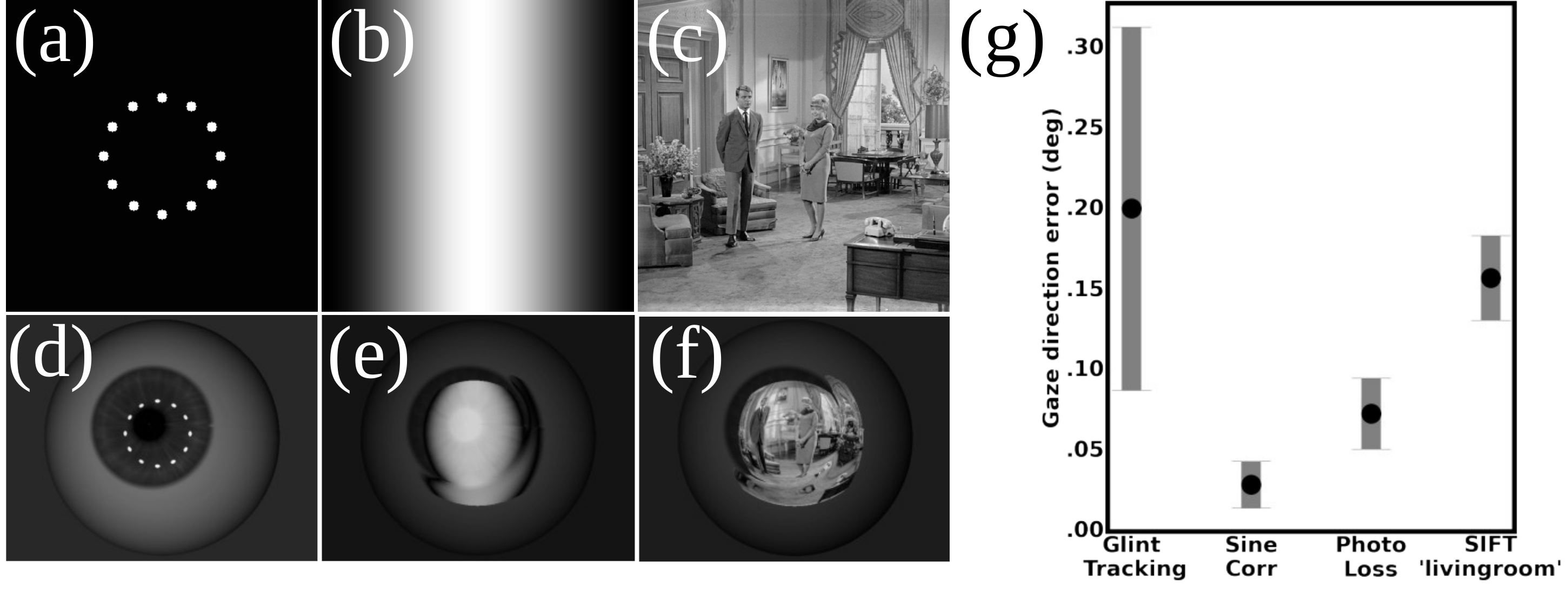}
    \caption{\textbf{Comparison of our different methods with glint tracking.} (a) Simulated glint pattern, used for our implementation of \cite{mestre2018robust} (b) Sinusoid pattern used with our correspondence  loss and photometric loss method (c) Living room image used with our correspondence loss method. (d)-(f) respective rendered images of the CG eye model. (g) Results and comparison. Correspondence loss performs the best in terms of average gaze direction error. }
    \label{disp}
\end{figure*}

In this section we compare our differentiable rendering optimization approach to state-of-the-art glint tracking gaze estimation techniques. Particularly, we follow the representative method by Mestre \etal \cite{mestre2018robust}, which uses  an interpolation based technique that utilizes the  pupil center and a ring of 12 glint illumination sources. We use our simulation pipeline to generate images of our CG eye model with corneal glint reflections (see Fig.~\ref{disp}d). Moreover, we implement the algorithmic steps described in \cite{mestre2018robust} to evaluate the gaze direction for each of our generated images. We compare our implementation of  \cite{mestre2018robust} against our developed  optimization  frameworks in simulation using either  correspondence loss (Sec.~\ref{corr}), or photometric loss (Sec.~\ref{col}). For both of our methods, we use a sinusoid pattern with 1 period in horizontal screen direction (see Figs.~\ref{disp}b and e). For the correspondence loss method, we phaseshift the sinusoid to extract the correspondences, while we only use one sinusoid image (single-shot!) for the photometric loss method. Additionally, we perform a simulated experiment where we display an arbitrary image (``living room''), which is representative for  movie content or a video game stream that is displayed on the VR/AR/MR headset (see Figs.~\ref{disp}c and f). We use the top 50 closest SIFT feature matches between the ``living room'' image (Fig.~\ref{disp}c) and the rendered eye image (Fig.~\ref{disp}f) to extract image-screen correspondences for our correspondence-loss method. As we only use one image per gaze direction evaluation, this procedure is also single-shot.

One important restriction of the implementation of the glint tracking method is that the point source reflections always need to ``stay'' on the cornea, which restricts the allowed movement of the eye. We therefore limited our eye movement to rotations only and set the translations to 0 for this experiment. We tested all methods under the same set of 50 random gaze angles with the elevation and azimuth of the gaze both under $\pm 5$ degrees. For all simulated images, we applied an additional 5 percent Poisson noise to simulate camera shot noise. 

Our results are summarized in Fig.~\ref{disp}g: for each of the different methods, we calculated the mean error of all measurements with respect to the ground truth gazing directions (shown as point), as well as the precision (shown as bar). It can be seen that, compared to the glint tracking implementation,  our correlation loss and photometric loss methods achieve a much lower average error in eye gaze direction estimation and much higher precision. We achieve a average gaze direction error of $0.03^\circ$ for the correlation loss and $0.07^\circ$ photometric loss, compared to $0.20^\circ$ for our glint tracking implementation. This demonstrates that a low gazing evaluation error can be achieved by (1) additional dense information from an extended light source (compared to discrete glint illumination), and (2) by explicitly utilizing the 3D modeling information in the 3D scene. We also note that the ``living room'' experiment using SIFT features  achieved competitive performance as well, with an average gaze direction error that is still better than the gaze error for glint tracking ($0.15^\circ$ vs. $0.20^\circ$) and a much better precision. Additionally, we note that our simulated result of the glint tracking method is better than the results specified in the original paper by Mestre \etal \cite{mestre2018robust}. This could be caused by the absence of a periocular region (eyelids, lashes, etc.) in simulation that could occlude the eye, as well as by a slight mismatch in noise levels between simulation and real experiment.

\subsection{Experimenting with increasingly sparser correspondences}
To further quantify the claimed advantage or a dense deflectometric surface information
 we conduct an additional real world experiment, similar to the experiment described in Sec.~\ref{real_exp}. The details of the experiment (used angles, number of measurements, etc.) are specified in the Supp. Material. We run our pipeline several times while \textit{artificially thinning out} the measured correspondence points in a random fashion, resulting in real world measurements with increasingly sparser correspondences. Our results in Tab.~\ref{tab:sparse_tb}  show that the dataset with the full correspondence delivers the best result in terms of both precision and mean error, and that the performance decreases with decreasing number of correspondence points. A further discussion of this experiment, in particular in comparison with glint tracking, is given in the Supp. Material.

\begin{table}[!htp]
\center
\caption{\textbf{Real world experiment for increasingly sparser amount of correspondences}}
\label{tab:sparse_tb}
\scriptsize
\begin{tabular}{l|r|r|r|r|r}\toprule 
Completeness &Full &1/50 &1/100 & 1/200 &1/500 \\ \midrule
Precision   &0.19$^\circ$ &0.23$^\circ$ &0.28$^\circ$ & 0.54$^\circ$ &4.52$^\circ$ \\
Mean error  &0.42$^\circ$ &0.63$^\circ$ &0.51$^\circ$ & 0.71$^\circ$ &1.52$^\circ$ \\

\bottomrule

\end{tabular}
\vspace{-4mm}
\end{table}

Besides the \textit{number} of correspondences, the \textit{quality} of correspondences is estimated to have a significant effect on the gaze estimation error as well. Here, the selection of the pattern plays a crucial role. For example, it is generally know from metrology literature that a high frequency sinusoid pattern tends to be more noise robust and generally delivers better correspondences than low frequency sinusoids. A study of this effect is shown in the Supp. Material.

\section{Discussion and Conclusion}
In this paper we introduced a novel optimization-based eye tracking framework that exploits extended screen illumination instead of sparse single point light sources. We developed a differentiable pipeline that accurately simulates specular reflection and extracts useful 3D information to guide our optimization, while jointly optimizing for the shape of our eye model. We conducted quantitative real-world experiments and demonstrated that our method performs better than other representative reflection based methods, meaning that the dense correspondence information we obtain from screen illumination is  crucial to accurate eye tracking.

However, our introduced method is not without limitations. First, our shape optimization process is still tied to a very specific model. Other variation parameters such as sphere radius or distance of sphere centers could be explored and might possibly even lead to a simplification. On the other hand, the optimization operates under the assumption of rotational invariance around the gaze axis, ignoring potential asymmetrical deformations of the cornea and sclera. To address this issue, we  could refine our optimization algorithm in the future.  For example, we can consider a coarse-to-fine optimization approach for our shape optimization pipeline to introduce local deformations. Moreover, we could potentially improve our photometric loss method by  jointly optimizing the reflectance map of the eye along with the periocular region. Lastly, our experiments only use one single camera view to perform our optimization. We hypothesize that our results can be further improved by employing a stereo or multi-view setup, as this would add coverage to the reflection area of the eye -- however at the cost of a higher system complexity. 

We believe that our novel approach to eye tracking could potentially usher a new wave of gaze estimation principles that could lead to faster, less hardware-intensive, and more accurate, eye tracking in the future.

{\small
\bibliographystyle{ieee_fullname}
\bibliography{Arxiv-Submission_V1}
}

\end{document}